\documentclass[conference]{IEEEtran}
\ifCLASSINFOpdf
   \usepackage[pdftex]{graphicx}
   \graphicspath{{./figures/}{../jpeg/}}
   \DeclareGraphicsExtensions{.pdf,.jpeg,.png}
\else
   \usepackage[dvips]{graphicx}
   \graphicspath{{figures/}}
   \DeclareGraphicsExtensions{.eps}
\fi

\usepackage{booktabs}
\usepackage{todonotes}
\usepackage{import}
\usepackage{xcolor}
\usepackage{units}
\usepackage{textcomp}
\usepackage{float}
\usepackage[nolist,nohyperlinks]{acronym}
\definecolor{green}{RGB}{44,160,44}
\definecolor{red}{RGB}{214,39,40}
\begin{document}
%
\title{Multi-Label Wireless Interference Identification with Convolutional Neural Networks}

\author{\IEEEauthorblockN{Sergej Grunau, Dimitri Block, Uwe Meier}
\IEEEauthorblockA{inIT - Institute Industrial IT\\
OWL  University of Applied Sciences\\
Lemgo, Germany\\
\{sergej.grunau, dimitri.block, uwe.meier\}@hs-owl.de}}


%


\maketitle

\begin{abstract}
The steadily growing use of license-free frequency bands requires reliable coexistence management and therefore proper \ac{WII}.
In this work, we propose a \ac{WII} approach based upon a deep
\ac{CNN} which classifies multiple IEEE~802.15.1, IEEE~802.11 b/g and IEEE~802.15.4 interfering signals in the presence of a utilized signal.
The generated multi-label dataset contains frequency- and time-limited sensing snapshots with the bandwidth of \unit[10]{MHz} and duration of \unit[12.8]{\textmu s}, respectively.
Each snapshot combines one utilized signal with up to multiple interfering signals.

The approach shows promising results for \acl{STI} with a classification accuracy of approximately \unit[100]{\%} for IEEE~802.15.1 and IEEE~802.15.4 signals. 
For IEEE~802.11 b/g signals the accuracy increases for \acl{CTI} with at least \unit[90]{\%}. 
\end{abstract}


\IEEEpeerreviewmaketitle

\acresetall
\section{Introduction}\label{sec:introduction}

Artificial neural networks and especially \acp{CNN} achieved excellent results for different benchmarks in recent years~\cite{2012.CiresanMeierSchmidhuber,2012.KrizhevskySutskeverHinton,2013.LinChenYan}.
Neural networks achieve the best performance, e. g. for character recognition of the \ac{MNIST} database~\cite{2016.LecunCortes}. 
The results achieved by the \acp{CNN} from Cire\c{s}an et al.~\cite{2012.CiresanMeierSchmidhuber} are comparable to human performance. 
Therefore, a growing number of research fields apply \acp{CNN} as classification systems.

One of these research fields is \ac{WII} for coexistence management of license-free radio bands such as the \unit[2.4]{GHz} \ac{ISM} band.
Such bands are shared between incompatible heterogeneous wireless \acp{WCS}. 
In industrial environments, typically standardized \acp{WT} within the \unit[2.4]{GHz} \ac{ISM} band are wide-band high-rate IEEE 802.11b/g/n, narrow-band low-rate IEEE 802.15.4-based WirelessHART and ISA 100.11a, 
and IEEE 802.15.1-related PNO WSAN-FA and Bluetooth. 
Additionally, the radio band is shared with many proprietary \acp{WT} which target specific application requirements such as industrial WLAN (iWLAN) from Siemens AG which is derived from IEEE 802.11, FHSS-based Trusted Wireless from Phoenix Contact and IEEE 802.15.1-based WISA from ABB Group.

Heterogeneous temporal and spectral medium utilization of shared radio bands result in interferences.
Any interference can cause packet loss and transmission latency for 
industrial \acp{WCS}.
Both effects have to be mitigated for real-time medium requirements.
The IEC 62657-2 norm \cite{2013.IEC626572:2013} for industrial \acp{WCS} recommends an active coexistence management for reliable medium utilization.
Therefore, it advises the utilization of (i) manual, (ii) automatic 
non-cooperative or (iii) automatic cooperative coexistence management.
The first approach is the most inefficient one, due to time-consuming configuration effort.
The automatic approaches (ii) and (iii) enable efficient self-reconfiguration without manual intervention and radio-specific expertise.
Automatic cooperative coexistence management (iii) requires a control channel, i. e. a logical common communication connection between each coexisting \ac{WCS} to enable deterministic medium access.
For legacy \acp{WCS} without control channel, the non-cooperative approach (ii) is recommended.
Non-cooperative coexistence management approaches are aware of coexisting \acp{WCS} based on independent \ac{WII} and mitigation.

\begin{figure}[t]
	\centering
	\includegraphics[width=0.99\linewidth]{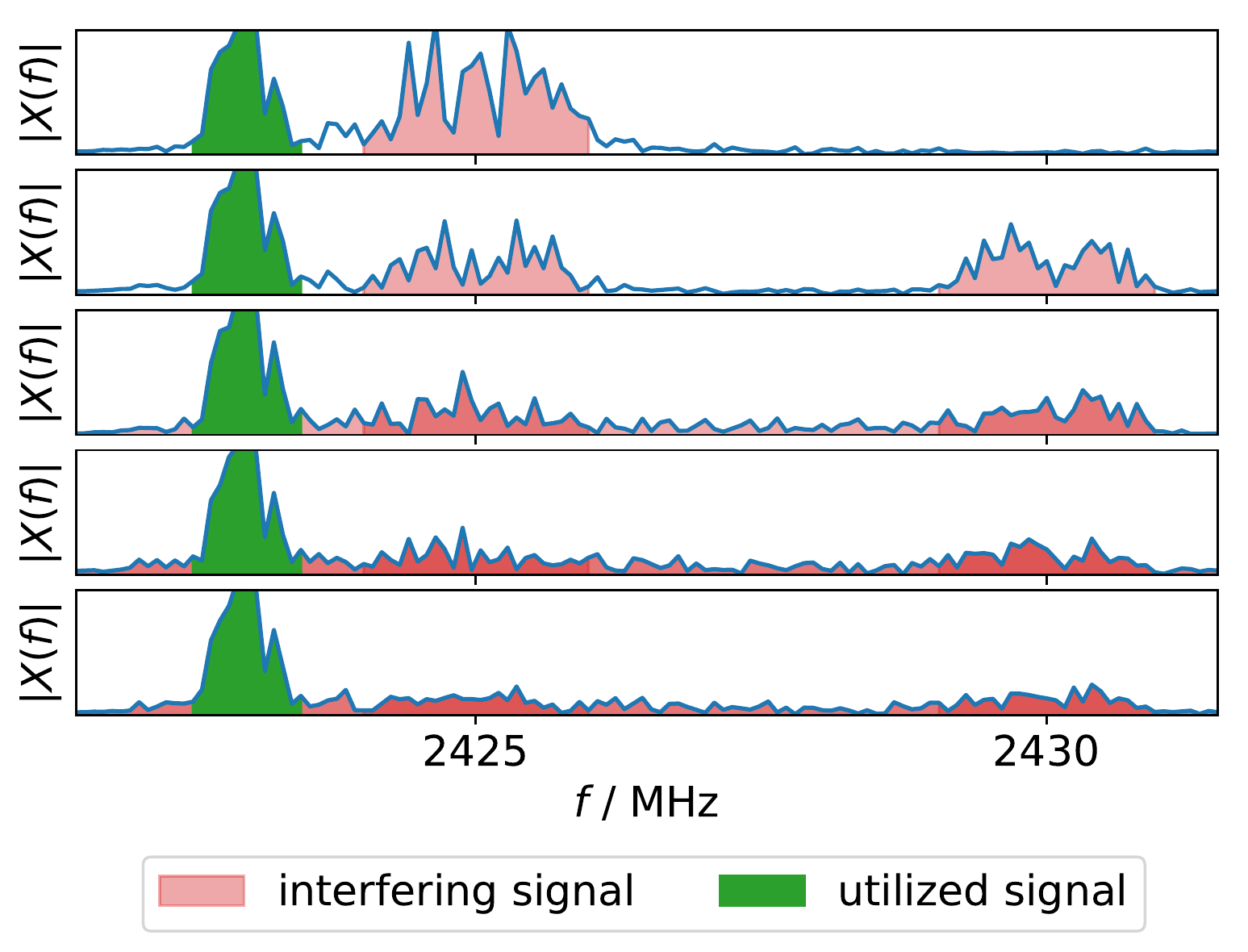}
	\caption{Exemplary \acl{CTI} snapshots with a utilized signal based on the \ac{WT} IEEE~802.15.1 and an increasing number of interfering signals based on the \acp{WT} IEEE~802.15.4 and IEEE~802.11 b/g}
	\label{fig:multi_label_snapshots_cti}
\end{figure}

In our previous work \cite{Schmidt}, we proposed a \ac{WII} approach upon deep \acp{CNN}.
It classifies interference signals on a per sensing snapshot basis.
In this paper, we extend the approach to target coexistence management for an utilized \ac{WCS} which is interfered by non-cooperative \acp{WCS}.
Therefore, the proposed approach is capable of identification of multiple interfering signals in the presence of the utilized signal as illustrated in Fig.~\ref{fig:multi_label_snapshots_cti}.

To face realistic \ac{WCS} capabilities the approach is limited to a sensing bandwidth of \unit[10]{MHz}
and a sensing snapshot is limited to a duration of \unit[12.8]{\textmu s}, which results in 128 \ac{IQ} samples.
The evaluation is performed with the standardized \acp{WT} IEEE 802.11b/g, IEEE 802.15.4 and IEEE 802.15.1, 
which are sharing the \unit[2.4]{GHz} \ac{ISM} band. 
In total 19 different variants of modulation types and symbol rates are utilized.
Thereby, the \ac{WII} approach has to differ between 15 allocated frequency channels of the \acp{WT}.
Additionally, the sensing snapshots contain the dominant signal of the utilized \ac{WCS} which aggravates \ac{WII} even more.
Hence, the interfering signal is acquired in the presence of the utilized signal.
Hence, each sensing snapshot superimposes one utilized signal with multiple interfering signals.

The paper is structured as follows.
In the next chapter~\ref{sec:related_work}, the related work is discussed. 
Then, in chapter~\ref{sec:dataset}, the generated dataset is explained.
Chapter~\ref{sec:cnn_network_design} aims the \ac{CNN} design. 
Chapter~\ref{sec:results} shows the performance of the proposed approach. 
Finally, chapter~\ref{sec:conclusion} concludes the paper and suggests future work.

\begin{figure*}[tb]
	\centering
	\includegraphics[width=0.99\linewidth]{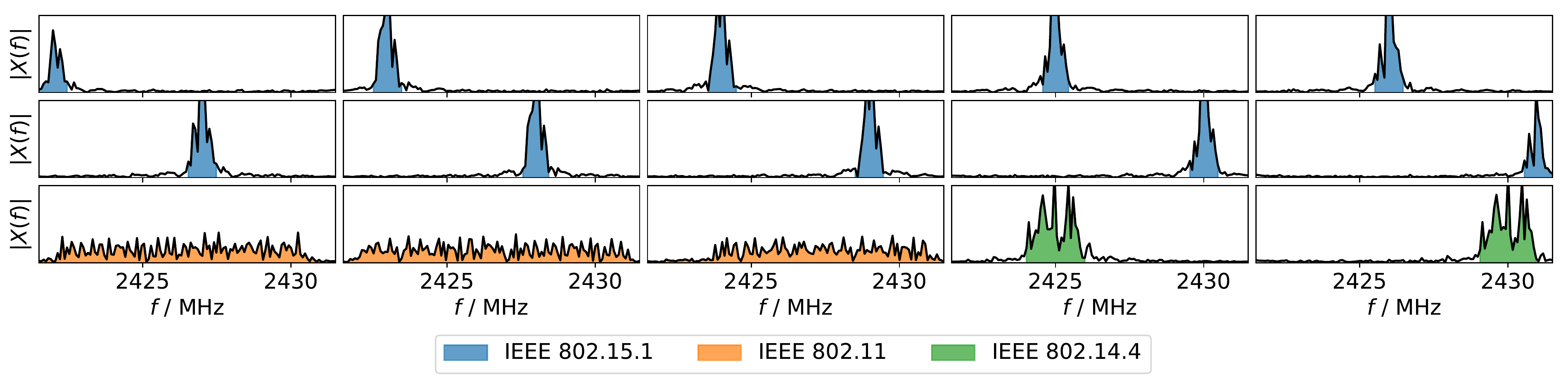}
	\caption{Snapshot examples from the single-label dataset of each class of the \aclp{WT} IEEE~802.15.1, IEEE~802.11 b/g and IEEE~802.15.4}
	\label{fig:single_label_example_snapshots}
\end{figure*}
\begin{figure}[H]
	\centering
	\includegraphics[width=0.9\linewidth]{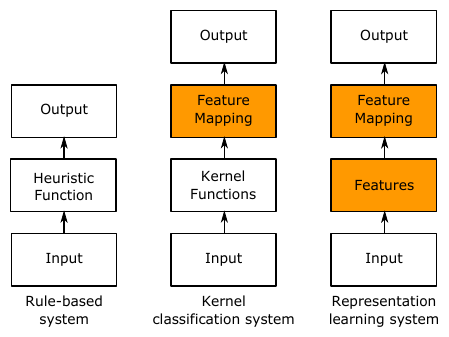}
	\caption{Classification system types with engineering (white) and self-optimizing (yellow) processing units}
	\label{grp_system}
\end{figure}

\section{Related Work}\label{sec:related_work}

Classification systems can be divided into (i) rule-based, (ii) kernel-based and (iii) representation learning ones as illustrated in Fig.~\ref{grp_system}.
Rule-based systems are mostly engineered implementations of heuristic functions to result in approximate solutions based on problem domain knowledge. 
In contrast, kernel classification systems self-optimize the weighting and assignment of pre-engineered features which are also based on problem domain knowledge. 
The finally classification system type is based on representation learning.
Thereby, also the feature extraction is performed by self-optimization.
Hence, representation learning systems eliminate any problem domain knowledge requirement and result in a full self-optimization classification system.

A well-known sub-type is called \ac{DNN}, which is a multi-layer neural network for higher-order feature extraction.
Another sub-type is a \ac{CNN}, which utilizes convolution-related feature extraction layers.
Hence, a deep \ac{CNN} classifies with higher-order convolution-related features.

The self-optimization processing units in classification systems require a training phase. 
For the training phase, these systems require a set of input data, which contain specific objects, and the output labels of the object classes. 
Thereby, the data type, objects, and the labels depend on the application domain.
Hence, classification systems assign input data objects to the label of the corresponding class.
With multiple classes, classification problems can be divided into (i) single- and (ii) multi-label ones. 
In the former, only one out of several objects is present in the input data.
Multi-label classification permits more than one object within the input data.

Within the problem domain of \ac{WII}, an input data item is a sensing snapshot.
Further, a contained object is a superimposed transmitted signal of a specific \ac{WCS}.
The particular frequency channel and \ac{WT} of the transmitted signal express the object classes and therefore also the labels.
The transmitted signals are distorted by the radio channel, e. g. attenuation and additive noise.
Additionally, superposition of multiple signals impairs the classification task.
Hence, \ac{WII} in the presence of a utilized signal raises the multi-label requirement.

\subsection{Neuro-Fuzzy Signal Classifier}
The first system is called \ac{NFSC}.
The \ac{NFSC} is a rule-based system that was implemented and tested in \cite{Block}. 
The \ac{NFSC} classifies frequency channels of IEEE~802.11 and IEEE~802.15.1 signals in six different industrial scenarios. 
The \ac{NFSC} consists of six layers. 
In these six layers, the \ac{NFSC} extracts feature from the input data.  
The extracted features are the center frequency, bandwidth, spectral pulse shape, time behavior, and spectral hopping behavior of the signals. 
Then, the features are assigned to the corresponding \ac{WT} and frequency channel class.
The \ac{NFSC} can detect multiple signals in the input data and can, therefore, handle a multi-label classification problem.
Nevertheless, the manual engineered heuristic functions lead to sub-optimal classification accuracies.

\subsection{Convolutional Neural Network Classification}
O'Shea et al. \cite{OShea} introduce an approach based on \acp{CNN}.
To recognize signal modulation types instead of \ac{WT} or frequency channels. 
The approach differentiates between eleven modulation types.
Thereby, the input data only contains the signal, and therefore it is a single-label classification problem. 
To train these system, a dataset with 96,000 snapshots was used.  
For radio channel distortion, a variable-strength white Gaussian noise was added to the dataset.
So, it consist of snapshots with a \ac{SNR} of \unit[-20]{dB} to \unit[20]{dB}. 
They show that \acs{CNN}-based self-optimizing systems outperform rule-based systems and also other \ac{DNN}-based systems.
\acs{CNN}-based systems achieve the best classification accuracies of greater than \unit[90]{\%} for a \ac{SNR} of at least \unit[-2]{dB}.

In our previous work, Schmidt et al. \cite{Schmidt} transferred the modulation recognition approach from O'Shea et al. to the problem domain of \ac{WII}.
The dataset was created using a \ac{VSG} for signal generation and a \ac{RSA} for data acquisition.
The sensing bandwidth was limited to a bandwidth of \unit[10]{MHz} and duration of \unit[12.8]{\textmu s}.
The dataset includes signals from the \acp{WT} IEEE~802.11~b/g, IEEE~802.15.1, and IEEE~802.15.4 with there in-band frequency channels.
They are divided into fifteen classes. 

Schmidt et al. evaluated two different network architectures of \acp{CNN}. 
The first architecture was adapted from O'Shea et al. \cite{OShea}, while the latter one was a reduced variant to avoid overfitting.
However the classification accuracy of the first architecture is higher during the training phase, there is hardly any difference in the validation phase.
It results in classification accuracies of at least \unit[95]{\%} at \ac{SNR} of \unit[-5]{dB}.
Additionally, the \acp{CNN} outperforms the \ac{NFSC} regarding classification accuracy.
It shows a processing gain of \unit[5.32]{dB} and a classification accuracy improvement of \unit[8.19]{\%}.
But the \acp{CNN} is limited due to its single-label capability only to classify one signal from each sensing snapshot.

\section{Multi-Label Dataset Generation}
\label{sec:dataset}

The multi-label \ac{WII} has to detect interference signals in the presence of a utilized signal of the \acp{WT} IEEE~802.15.1, IEEE~802.11 b/g, and IEEE~802.15.4. 
To approach real \ac{WCS} capabilities, a limited sensing bandwidth of \unit[10]{MHz} was assumed.
Hence, eight simultaneous operating instances are required for loss-less \ac{WII} within the \unit[2.4]{GHz} band. 
Each sensing band covers ten, three and two frequency channels of the \acp{WT} IEEE~802.15.1, IEEE~802.11 b/g, and IEEE~802.15.4, respectively. 
Thereby, the frequency channels of the \ac{WT}  IEEE~802.11 are only partially within the sensing bandwidth. 

\begin{figure}[tb]
	\centering
	\includegraphics[width=0.9\linewidth]{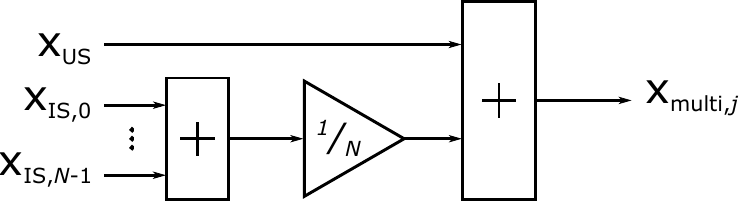}
	\caption{Multi-label snapshot generation based on single-label snapshots of a utilized signal $x_{\mathrm{US}}$ and a weighted sum of multiple interfering signals $x_{\mathrm{IS, i}}$}
	\label{fig:multi_label_dataset_generation}
\end{figure}

The training and validation dataset for the multi-label \ac{WII} was derived from the one of Schmidt et al. \cite{Schmidt} as illustrated in Fig~\ref{fig:multi_label_dataset_generation}. 
They generated a synthetic dataset $D_{\mathrm{single}}$ with \ac{VSG} stimulation and \ac{RSA} recording. 
$D_{\mathrm{single}}$ consists of several snapshots $x_{\mathrm{single}, i}$ and labels $Y_{\mathrm{single}, i}$. 
Each snapshot $x_{\mathrm{single}, i}$ is represented by complex 128 \ac{IQ} samples and contains one of fifteen different classes. 
The classes represent the ten, three and two frequency channels of the \ac{WT} IEEE~802.15.1, IEEE~802.11 b/g, and IEEE~802.15.4, respectively.
Additionally, $D_{\mathrm{single}}$ utilized \ac{AWGN} such that the \ac{SNR} varies between \unit[-20]{dB} up to \unit[20]{dB} with a step size of \unit[2]{dB}. 
In total, $D_{\mathrm{single}}$ contains 225,225 snapshots, with 715 snapshots per class and \ac{SNR} combination.

The multi-label dataset $D_{\mathrm{multi}}$ requires snapshots with several classes.
Thereby, one class is the utilized signal, and the remainders are interfering signals.
Since it is not likely that all fifteen classes occur in a snapshot simultaneously, it is limited to the utilized signal an up to six interfering signals.
Additionally, despite the varying number of interfering signals the \ac{SIR} remains constant.
Hence, the results depend only on the amount of interfering signals.
Furthermore, classification of an increasing number of interfering signals with a fixed \ac{SIR} is more challenging.

\begin{figure}[tb]
	\centering
	\includegraphics[width=0.99\linewidth]{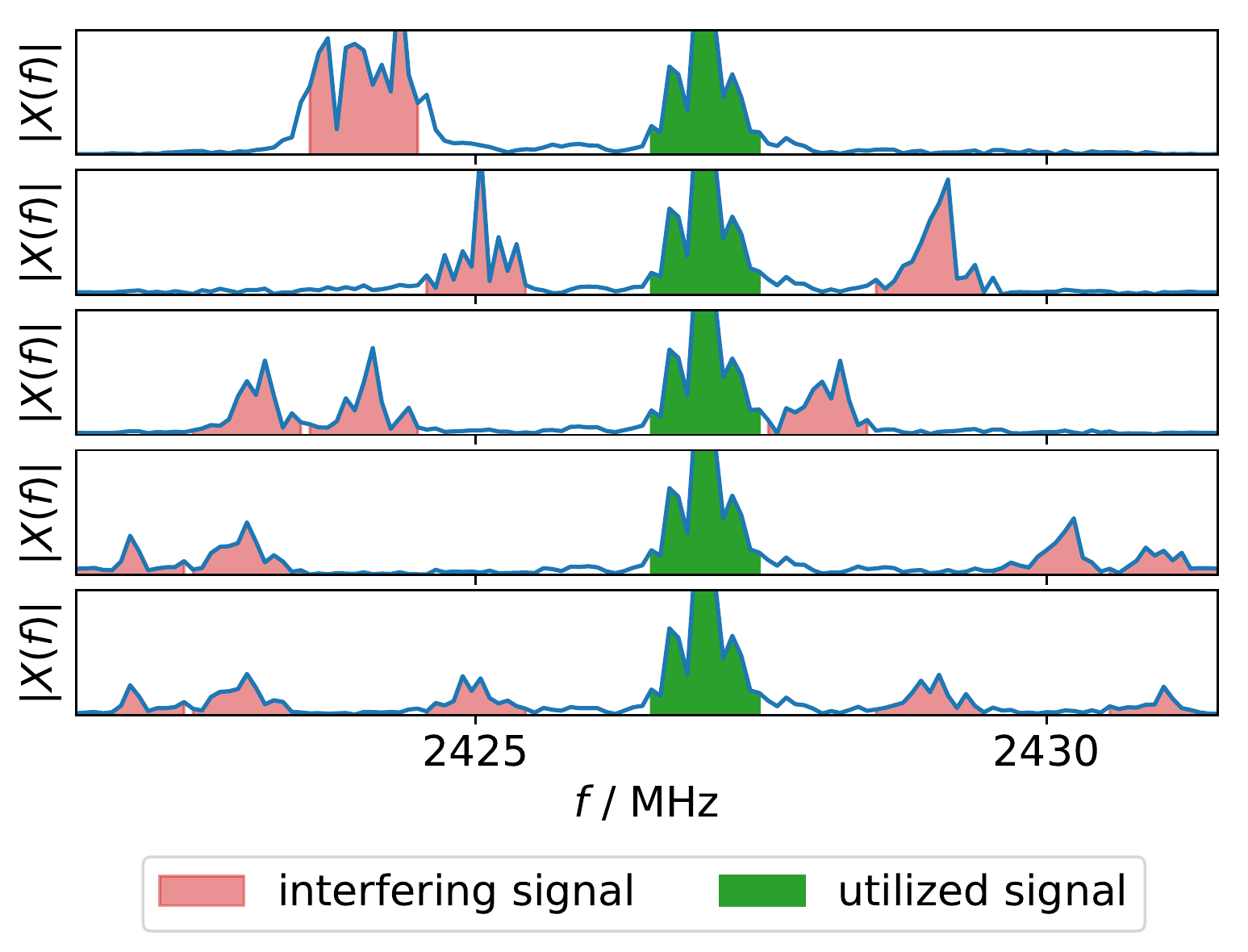}
	\caption{Exemplary \acl{STI} snapshots with a utilized signal and an increasing number of interfering signals with same \ac{WT} IEEE~802.15.1}
	\label{fig:multi_label_snapshots_sti}
\end{figure}

Fig.~\ref{fig:multi_label_dataset_generation} shows the multi-label generation signal flow with the combination of the single-label snapshots.
Each resulting snapshot $x_{\mathrm{multi}, j}$ contains a single utilized signal and $N$ interfering signals.
Hence, the corresponding label is the union of the labels of the contained classes.
For input single-label snapshots with a \ac{SNR} of \unit[20]{dB} are used.
Furthermore, the interfering signals were weighted with the factor $1/N$ to keep the \ac{SIR} constant with the value one. 
Resulting same- and \acl{CTI} (\acsu{STI}, \acsu{CTI}) snapshots are illustrated in Fig.~\ref{fig:multi_label_snapshots_sti} and Fig.~\ref{fig:multi_label_snapshots_cti}, respectivly.


The entire dataset $D_{\mathrm{multi}}$ consists of 450,000 snapshots and labels. 
These are divided into 360,000 snapshots for training and 90,000 snapshots for validation purposes.

\section{Neural Network Design}
\label{sec:cnn_network_design}

The \ac{CNN} utilizes some pre-processing for the input data. 
Schmidt et al. \cite{Schmidt} and Danev et al. \cite{Danev} have shown that the classification of the frequency representation of radio signals increases the classification accuracy.
Therefore, the snapshots are transformed with the \ac{FFT}. 
Then, the resulting 128 complex values have been translated into a $128 \times 2$ matrix with the extracted real and imaginary parts.
Thereby, real values are in the first column and the imaginary values in the second one. 

The \ac{CNN} output is a vector with 15 elements with the value range $[0, 1]$. 
Thereby, each element represents a class. 
For validation, a threshold is applied to result in binary output.

\subsection{Network Architecture}

The network architecture of the \ac{CNN} is shown in Tab.~\ref{tab:cnn_struc}. 
It is derived from the \ac{CNN} architectures of Schmidt et al. \cite{Schmidt} and O'Shea et al. \cite{OShea}. 
Thereby, Schmidt et al. have used a softmax activation function at the output of the last layer for optimal single-label classification. 
However, softmax activation function is not suitable for a multi-label classification problem.
Therefore, it has been replaced by a sigmoid activation function.
The sigmoid activation function enables the independent output calculation for each a class.

\begin{table}[h]
\caption{CNN Architecture}
\label{tab:cnn_struc}
\centering
\begin{tabular}{llll}
    \toprule
    \textbf{Layer type} & \textbf{Input size} & \textbf{Parameters} & \textbf{Act. fct.}\\
    \toprule
    Convolutional & $128 \times 2$ & $3 \times 1$ filter kernel & Rectified \\
    layer & & $64$ feature maps & linear \\
    \midrule
    Convolutional & $64 \times 126 \times 2$ & $3 \times 1$ filter kernel & Rectified \\
    layer & & $1024$ feature maps & \\
        & & Dropout \unit[60]{\%} & linear \\ 
    \midrule
    Dense layer & $126,976 \times 1$ & $128 $ neurons& Rectified \\
        & & Dropout \unit[60]{\%} & linear \\
    \midrule
    Dense layer & $128 \times 1$ & $15$ neurons& Sigmoid\\
    \bottomrule
\end{tabular}
\end{table}

\subsection{Network Training}

The \ac{CNN} was trained in 200 epochs. 
The Adam optimization was used with the standard default parameters and a learning rate of 0.001 \cite{Kingma}. 
As a cost function, the binary cross entropy was used as which is the optimal choice for sigmoid output activation functions. 
The batch size of 256 has been adjusted to the limitations of the computing platform. 
Additionally, no hyperparameter optimization was applied.

\subsection{Implementation Aspects}

The \ac{CNN} was implemented in the programming language Python with the libraries Keras \cite{keras} and Tensorflow \cite{tensorflow}. 
A high end platform with an Intel XENON E5-1660 v3 \ac{CPU}, 16 GB RAM and a Nvidia GTX 960 \ac{GPU} was used. 
During the training process a \ac{CNN}, an epoch took \unit[390]{s}, resulting in a duration of \unit[21.6]{h} for training.

\section{Results}
\label{sec:results}

\begin{figure}[H]
	\centering
	\includegraphics[width=0.99\linewidth]{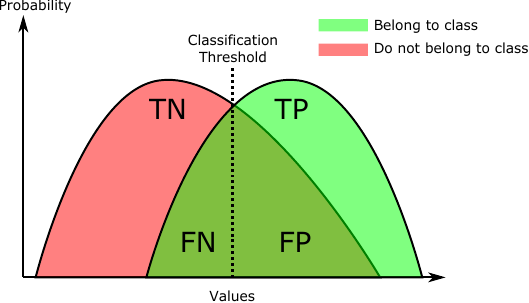}
	\caption{Evaluation metric \acl{TPR} $TPR = TP / (TP + FN)$  expresses the proportion of the correct classified data items out of all items belonging to the certain class}
	\label{fig:tpr}
\end{figure}

For evaluation, we use a metric called \ac{TPR}, which is proposed by Godbole and Sarawagi \cite{Godbole}.
\ac{TPR} evaluates the outcome of a single class.
Thereby, all data items which contain the distinct class are considered.
\ac{TPR} expresses the proportion of the actual correct classified ones, as illustrated in Fig.\ref{fig:tpr}.

It is important to note, that \ac{WII} targets the classification of the interfering signals.
The known utilized signal is, therefore, an unwanted one and therefore distorts the snapshots.

\subsection{Single-Label Classification}

\begin{figure}[H]
	\centering
	\includegraphics[width=0.99\linewidth]{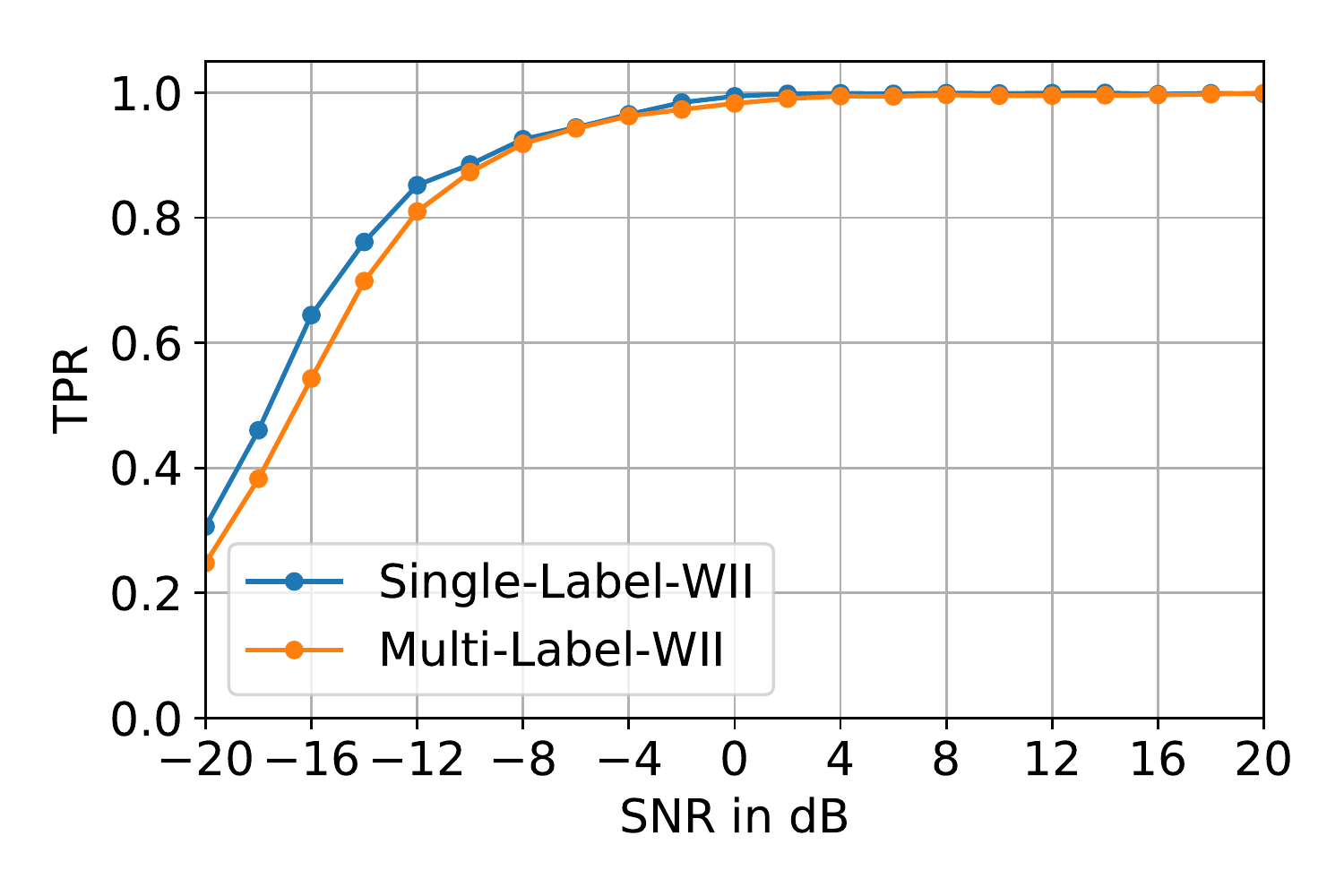}
	\caption{Single-label classification comparison of the proposed multi-label \ac{WII} with the single-label one from Schmidt et al. \cite{Schmidt}}
	\label{fig:single_vs_multi_wii}
\end{figure}

First, the proposed multi-label \ac{WII} approach was compared with single-label \ac{WII} one from Schmidt et al. \cite{Schmidt}.
Therefore, the same single-label validation dataset from Schmidt et al. was utilized.
The resulting averaged \ac{TPR} is shown in Fig.~\ref{fig:single_vs_multi_wii} with the varying \ac{SNR}.
Thereby, the proposed approach only differs for a \ac{SNR} below \unit[-8]{dB}.
Below, the multi-label \ac{WII} approach is up to \unit[10.14]{\%} worse and subtracts a processing gain of up to \unit[1.1]{dB}.
Hence, the single-label \ac{WII} approach from Schmidt et al. \cite{Schmidt} results in slightly better performance with the assumption of single-label data, and therefore without any utilized signal.

\subsection{Same-Technology Interference Classification}

\begin{figure}[H]
	\centering
	\includegraphics[width=0.99\linewidth]{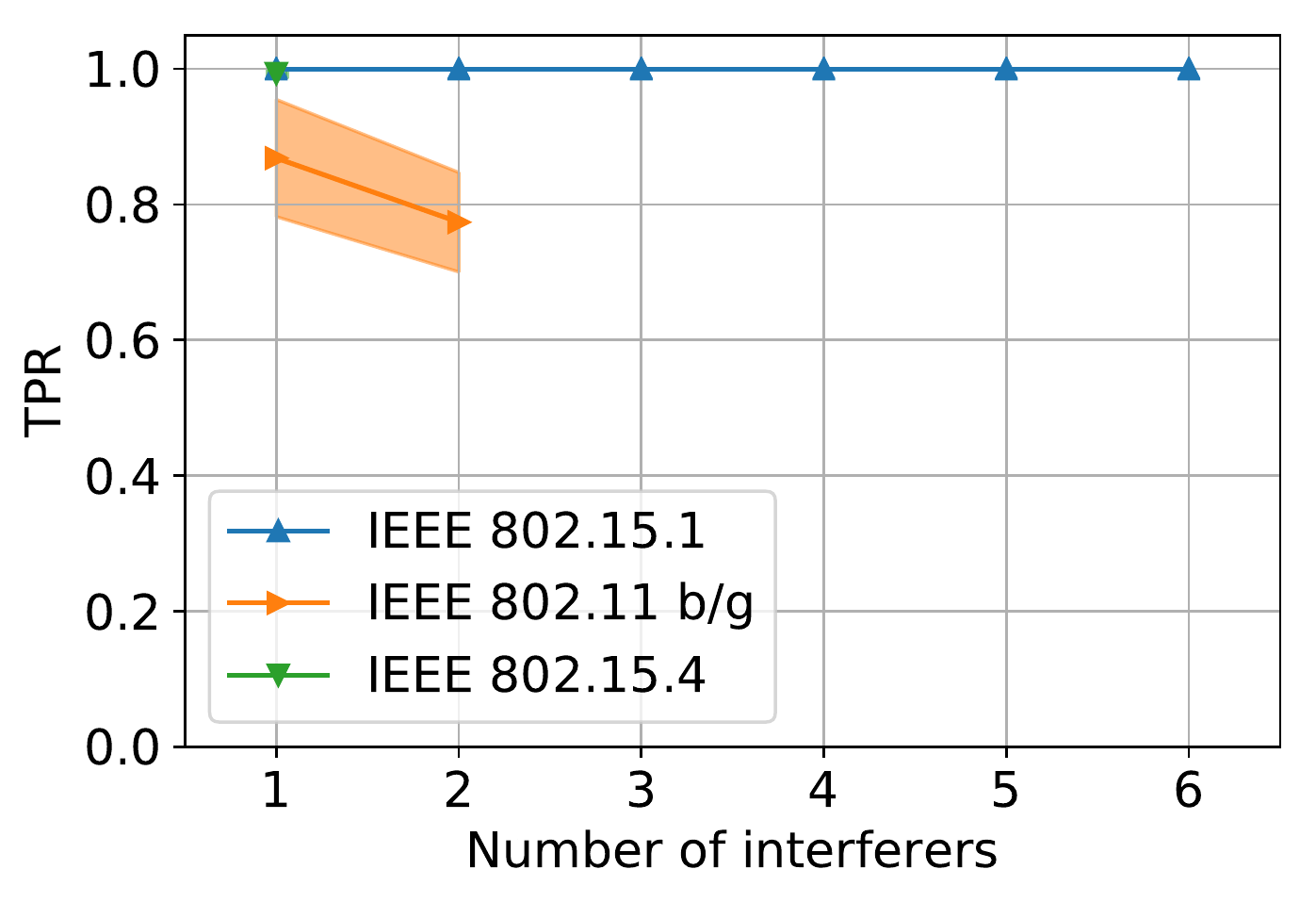}
	\caption{Same-technology interference \ac{WII} with the mean \ac{TPR} (solid line) and the 10\% / 90\% percentile interval (transparent area)}
	\label{fig:res_sti}
\end{figure}

The results of the multi-label \ac{WII} approach with \ac{STI} are shown in Fig.~\ref{fig:res_sti}. 
Thereby, the snapshots contain a utilized signal and a varying number of interfering signals of the same \ac{WT}.

The classification of \acs{STI} with the narrow-band \acp{WT} IEEE~802.15.1 and IEEE~802.15.4 approach a \ac{TPR} of one.
The optimal behavior results from the limited spectral overlapping of the superimposed signals.
Another reason is that the sensing bandwidth entirely covers signals from both \acp{WT} IEEE~802.15.1 and IEEE~802.15.4.
Additionally, the \ac{TPR} of IEEE~802.15.1 \ac{STI} interferences is independent of the number of interfering signals, and therefore also it is independent of the frequency channel of the utilized signal.

In contrast, the \ac{TPR} of a wide-band IEEE~802.11 b/g \ac{STI} signal is worse and drops even further with multiple signals.
Such signals are only partially within the sensing bandwidth of a snapshot. 
Another reason is that the inter-signal overlapping is significant.
Therefore, the differentiation between interfering signals and the utilized signal is more difficult. 
Additionally, the high variation of the \ac{TPR} indicates the frequency channel dependency of the utilized signal.

\subsection{Cross-Technology Interference Classification}

\begin{figure*}[tb]
    \centering
    \includegraphics[width=0.99\linewidth]{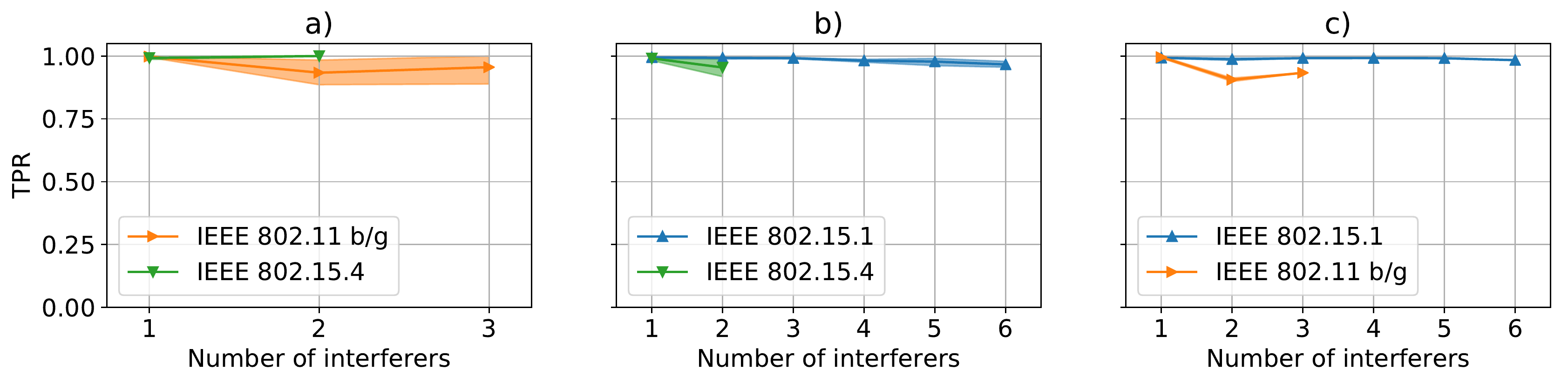}
    \caption{\ac{TPR} of the interfering signals with increasing number of signals for \ac{CTI}. Interference and  utilized signal are different technologies. The technology of the utilized signal is a) IEEE~802.15.1, b) IEEE~802.11 b/g and c) IEEE~802.15.4 with varying channels. The transparent area shows how the \ac{TPR} varies as the channel of the signal changes.}
    \label{fig:res_cti}
\end{figure*}

Figure~\ref{fig:res_cti} shows the evaluation for \acp{CTI}.
It is noticeable, that the \ac{TPR} for three IEEE~802.11 b/g interfering signals increases. 
This unexpected behavior may result from the combinatoric property that for three possible options the probability for one or three correct ones is higher than two correct ones. 
It is also possible that the \ac{CNN} is overfitting for three IEEE~802.11 b/g interferers. 

For \acs{CTI} with an IEEE~802.11 b/g utilized signal the proposed approach results in a slightly better \ac{TPR}. 
This is because the bandwidth of the IEEE~802.15.1 and IEEE~802.15.4 signals are narrow compared to the snapshot bandwidth and therefore less overlapping interference occurs.

In case of a wide-band IEEE~802.11 b/g utilized signal the narrow-band interfering signals suffer from the spectral intersection.
It results in a decrease of the \ac{TPR}.
Additionally, the high variation of the \ac{TPR} indicates the frequency channel dependency of the utilized signal.

\acresetall
\section{Conclusion}
\label{sec:conclusion}
The steadily growing use of license-free frequency bands requires reliable coexistence management and therefore proper \ac{WII}.
In this work, we propose a \ac{WII} approach based upon deep
\ac{CNN} which extends our work \cite{Schmidt}. The \ac{CNN} naively learns
its features through self-optimization during an extensive data-driven training process. In contrast to our previous work, we target coexistence management for cooperative utilized \ac{WCS} which are interfered by non-cooperative \acp{WCS}. 
We analyzed \acp{WCS} with the \acp{WT} IEEE~802.15.1, IEEE~802.11 b/g and IEEE~802.15.4.
Hence, our approach classifies multiple interfering signals in the presence of a utilized signal. 
Therefore, it is multi-class and multi-label classification problem.

For multi-label dataset generation, the single-label one form Schmidt et al. \cite{Schmidt} has been combined.
They are frequency- and time-limited with the bandwidth of \unit[10]{MHz} and duration of \unit[12.8]{\textmu s}, respectively.
Therefore, the \unit[2.4-]{GHz}-ISM band is divided into eight spectral non-overlapping sensing sub-bands.
Each sub-band contains fifteen frequency channels of the \acp{WT}, and therefore also fifteen classes.
The multi-label dataset contains in total 450,000 snapshots. 
Each snapshot combines one utilized signal with up to six interfering signals
with a \ac{SIR} of one.

The approach shows promising results for same- as well as for cross-technology interference (STI, CTI).
The \ac{STI} classification accuracies of spectral non-overlapping narrow-band IEEE~802.15.1 and IEEE~802.15.4 are approximately \unit[100]{\%}.
In contrast, spectral overlapping wide-band IEEE~802.11 b/g suffers from low accuracy of at least \unit[78]{\%}.

For \ac{CTI} IEEE~802.15.1 and IEEE~802.15.4, the accuracy slightly drops down to at least \unit[95]{\%}, because the utilized and interfering signals are partly spectral overlapping.
However, for IEEE~802.11 b/g the accuracy even increases with at least \unit[90]{\%}.
It takes advantage of the narrow-band utilized signal.

For future work, the evaluation has to be experimentally validated within industrial environments.

\section{Acknowledgement}
Part of this research was founded by KoMe (IGF 18350 BG/3 over DFAM, Germany) and HiFlecs (16KIS0266 over BMBF, Germany).

\bibliographystyle{IEEEtran}
\bibliography{literatur}
%
%
\begin{acronym}[Bash]
	\acro{CNN}{convolutional neural network}
	\acro{CPU}{central processing unit}
	\acro{CTI}{cross-technology interference}
	\acro{DNN}{deep neural network}
	\acro{FFT}{fast fourier transform}
	\acro{GPU}{graphical processing unit}
	\acro{GPU}{graphics processing unit}
	\acro{MNIST}{mixed National Institute of Standards and Technology}
	\acro{NFSC}{neuro-fuzzy signal classifier}
	\acro{RSA}{real time spectrum analyzer}
	\acro{SIR}{signal-to-interference ratio}
	\acro{SNR}{signal-to-noise ratio}
	\acro{STI}{same-technology interference}
	\acro{TPR}{true positiv rate}
	\acro{VSG}{vector signal generator}
	\acro{WCS}{wireless comunication system}
	\acro{WII}{wireless interference identification}
	\acro{WT}{wireless technology} \acroplural{WT}[WTs]{wireless technologies}
	\acro{ISM}{industrial, scientific and medical}
	\acro{IQ}{in-phase and quadrature}
	\acro{AWGN}{additive white Gaussian noise}
\end{acronym}

\end{document}